\documentclass[journal]{IAENGtran}
\usepackage{amssymb}
\usepackage{float}
\usepackage{ifpdf}
\usepackage{amsmath}

\usepackage{subfig}
\usepackage{cite}
\usepackage{xcolor} 
\usepackage{amssymb}

\ifCLASSINFOpdf
   \usepackage[pdftex]{graphicx}
   \DeclareGraphicsExtensions{.pdf,.jpeg,.png}
\else
   \usepackage[dvips]{graphicx}
   \DeclareGraphicsExtensions{.eps}
\fi

\usepackage{algorithm}
\usepackage{algpseudocode}
\usepackage{algorithmicx}
\usepackage[marginal]{footmisc}

\usepackage{array}

\usepackage{mdwmath}
\usepackage{mdwtab}
\usepackage{multirow} 

\usepackage{eqparbox}
\usepackage{fixltx2e}

\usepackage{stfloats}
\usepackage{hyperref}
\usepackage{url}

\begin{document}
%
\title{LayerTracer: A Joint Task-Particle and Vulnerable-Layer Analysis framework for Arbitrary Large Language Model Architectures}
%
%
%

\author{Yuhang Wu, Qinyuan Liu, Qiuyang Zhao, Qingwei Chong
\thanks{Yuhang Wu is a researcher of China Electronic Technology Nanhu Research Institute, Jiaxing, China}
\thanks{Qinyuan Liu is a researcher of China Electronic Technology Nanhu Research Institute, Jiaxing, China}
\thanks{Qiuyang Zhao is a researcher of China Electronic Technology Nanhu Research Institute, Jiaxing, China}
\thanks{Qingwei Chong is a deputy director of China Electronic Technology Nanhu Research Institute, Jiaxing, China (Corresponding Author)}
}

\maketitle
\pagestyle{empty}
\thispagestyle{empty}

\begin{abstract} 
Currently, Large Language Models (LLMs) feature a diversified architectural landscape, including traditional Transformer, GateDeltaNet, and Mamba. However, the evolutionary laws of hierarchical representations, task knowledge formation positions, and network robustness bottleneck mechanisms in various LLM architectures remain unclear, posing core challenges for hybrid architecture design and model optimization. This paper proposes LayerTracer, an architecture-agnostic end-to-end analysis framework compatible with any LLM architecture. By extracting hidden states layer-by-layer and mapping them to vocabulary probability distributions, it achieves joint analysis of task particle localization and layer vulnerability quantification. We define the task particle as the key layer where the target token probability first rises significantly , representing the model’s task execution starting point, and the vulnerable layer is defined as the layer with the maximum Jensen-Shannon (JS) divergence between output distributions before and after mask perturbation, reflecting its sensitivity to disturbances. Experiments on models of different parameter scales show that task particles mainly appear in the deep layers of the model regardless of parameter size, while larger-parameter models exhibit stronger hierarchical robustness. LayerTracer provides a scientific basis for layer division, module ratio, and gating switching of hybrid architectures, effectively optimizing model performance. It accurately locates task-effective layers and stability bottlenecks, offering universal support for LLM structure design and interpretability research.
\end{abstract}

\begin{IAENGkeywords}
Large Language Models; Task Particle; Vulnerable Layer; Hybrid Architecture; Jensen-Shannon
\end{IAENGkeywords}

%
\IAENGpeerreviewmaketitle

\section{Introduction}
\begin{figure}[!t]
    \centering
    \includegraphics[width=\columnwidth]{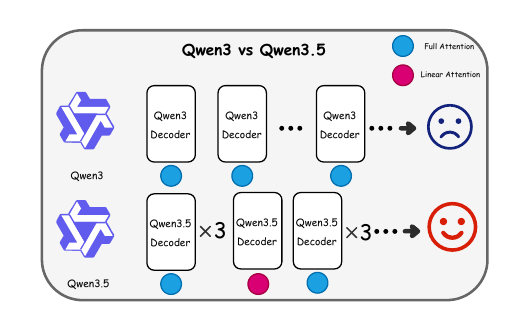}
    \caption{The architectures of Qwen3 model and Qwen3.5 model. Qwen3 adopts a single architecture, while Qwen3.5 is a hybrid architecture with a 3:1 ratio of Full Attention to Linear Attention}
    \label{fig:background}
\end{figure}

\IAENGPARstart{W}{ith} the rapid advancement of artificial intelligence technologies, Large Language Models (LLMs) have emerged as a transformative force. Representative models such as Qwen~\cite{IAENGhowto:qwen2tech,IAENGhowto:qwen2024,IAENGhowto:yang2025qwen3}, GPT~\cite{IAENGhowto:openai2023}, and LLaMA~\cite{IAENGhowto:touvron2023,IAENGhowto:touvron2023llama2,IAENGhowto:dubey2024llama3} have achieved breakthrough progress in natural language processing, intelligent interaction, and code generation by virtue of their powerful semantic understanding and reasoning capabilities. Moreover, these models effectively address the limitations of traditional approaches in long-text processing and complex logical reasoning, thereby significantly promoting the practical deployment of artificial intelligence technologies. Concurrently, model architectures have evolved beyond the conventional single-structure paradigm of the Transformer~\cite{IAENGhowto:vaswani2017}, and the combination of various architectures has become a mainstream design direction for balancing model performance with inference efficiency. For instance, novel architectures such as GatedDeltaNet~\cite{IAENGhowto:gateddelta2025}, Mamba~\cite{IAENGhowto:gu2024,IAENGhowto:dao2024}, and Linear Attention structures~\cite{IAENGhowto:ahn2024} continue to emerge, and their hybrid combinations have been widely adopted in model design. The most representative example is Qwen3.5~\cite{IAENGhowto:qwen35github}, a well-known hybrid architecture model that adopts a 3:1 ratio of Full Attention to Linear Attention. Specifically, three Decoder Layers with Full Attention are followed by one layer with Linear Attention in a cyclic manner. However, despite the remarkable achievements of such architectures, their ratio is essentially a black box and relies on intuition rather than scientific reasoning. As illustrated in Figure~\ref{fig:background}, the structural differences between Qwen3 (a single-structure model) and Qwen3.5 (a hybrid architecture model) are clearly presented, highlighting the typical characteristics of hybrid architecture design. In general, the architectural design of current large-scale models remains largely opaque~\cite{IAENGhowto:zhao2023}. In particular, the layer-wise allocation and module switching strategies for hybrid architectures lack scientific grounding. This situation hinders the optimization of architectural design and prevents the full exploitation of the core advantages offered by different architectures, representing a critical bottleneck that constrains both performance enhancement and engineering deployment of large-scale models.

Existing hierarchical analysis studies have applied probing approaches to investigate a wide range of tasks, yet they have not leveraged such techniques to analyze issues related to architectural design~\cite{IAENGhowto:ling2026neural}. First, the majority of these studies concentrate on the conventional Transformer architecture, conducting analyses solely on its attention mechanisms and hidden state characteristics~\cite{IAENGhowto:belinkov2022probing}, thereby failing to address the analytical requirements of emerging architectures such as GateDeltaNet~\cite{IAENGhowto:gateddelta2025} and Mamba~\cite{IAENGhowto:dao2024,IAENGhowto:gu2024}. Second, they focus exclusively on hierarchical representations within a single dimension and neglect to jointly model task-effective positions and network vulnerabilities~\cite{IAENGhowto:geva2021transformer}, rendering them inadequate for comprehensively reflecting the functional value of individual layers. Third, the absence of universal quantitative metrics for engineering implementation precludes direct guidance for layer partitioning and module allocation across diverse hybrid architectures~\cite{IAENGhowto:liu2023robustness}. Fourth, these studies fail to elucidate the intrinsic correlations among parameter scale, hierarchical knowledge utilization, and robustness~\cite{IAENGhowto:kaplan2020scaling}, thereby offering insufficient theoretical foundations for the scaling and optimization of large models with varying architectures.

To solve the above problems and break through the architecture adaptation limitations of traditional analysis frameworks as well as the shortcomings of existing studies, this paper proposes LayerTracer, a universal analysis framework. This framework adopts an architecture-agnostic design that enables seamless adaptation to any LLM architecture. It does not rely on core components of specific architectures, such as the attention head of Transformer, and can realize an integrated process of hierarchical behavior visualization, task particle automatic tracking and vulnerability quantitative analysis. We conduct joint analysis through two core quantifiable indicators. The first is task particle, which is defined as the key layer where the probability of the target word first rises significantly and continues to increase. It represents the starting position where the model begins to execute the target task. The second is vulnerable layer, which is defined as the layer with the maximum JS divergence of the output distribution before and after mask perturbation~\cite{IAENGhowto:lin1991}. It reflects the sensitivity of the network to perturbations. Through the joint analysis of these two core indicators, LayerTracer can accurately locate the task-effective layers and robustness bottlenecks of the model, thereby analyzing two core research questions: 
\begin{itemize} 
\item Why does Qwen3.5 adopt a 3:1 hybrid architecture ratio rather than 1:3?
\item What considerations should be taken when designing other hybrid architecture models?

\end{itemize}
This framework provides a scientific basis for the layer division, module ratio and switching strategy of various hybrid architectures, helps break the black box dilemma of LLM architecture design and supports the structural optimization and engineering implementation of large language models.


\section{Related Work}
\subsection{Layer-wise Interpretability Research of LLMs}
Existing research on the layer-wise interpretability of large LLMs mainly focuses on hierarchical representations, with two representative analytical paradigms: probing and non-probing methods. Probing methods insert additional probe modules into model layers to capture hidden state changes and locate task starting layers, but they often interfere with original layer representations and are limited to traditional Transformer architectures, failing to adapt to new architectures like GateDeltaNet and Mamba \cite{IAENGhowto:belinkov2022probing, IAENGhowto:gateddelta2025, IAENGhowto:gu2024}. Non-probing methods avoid such interference by directly extracting original hidden states to locate task-related layers, yet they only focus on positioning without extending results to robustness analysis or hybrid architecture design, and also lack adaptability to diverse architectures \cite{IAENGhowto:kim2023probing, IAENGhowto:lieber2024}.
In summary, both methods suffer from poor architecture adaptability and single-dimensional analysis. They fail to combine task starting layer positioning with robustness and architecture design needs, and non-probing methods further lack depth in exploring engineering value, failing to form guiding quantitative indicators or framework. These deficiencies hinder existing research from supporting the structural design, performance optimization and engineering implementation of diversified LLMs \cite{IAENGhowto:zhao2023, IAENGhowto:liu2024whatprobes}.

\subsection{Robustness and Vulnerability Analysis}
Research on model robustness mainly focuses on adversarial perturbations, input noise, and out-of-distribution generalization, using various indicators and methods such as gradient sensitivity, output distribution shift, and adversarial sample generation to identify model fragile components and robustness bottlenecks \cite{IAENGhowto:liu2023robustness}. Existing vulnerability analysis methods can be divided into overall analysis and layer-wise analysis, where overall analysis only evaluates the robustness of the entire model without accurately locating specific fragile layers, and layer-wise analysis, though capable of layer-level assessment, still has prominent shortcomings. Most layer-wise vulnerability analysis methods focus on the entire model or attention heads of the Transformer architecture, lacking fine-grained quantification of the overall vulnerability of each layer and relying heavily on unique components of the Transformer architecture, which makes them unable to adapt to non-Transformer architectures such as GateDeltaNet and Mamba \cite{IAENGhowto:gu2024, IAENGhowto:gateddelta2025}. Additionally, existing methods do not conduct joint analysis with task starting layer positioning, failing to explore the correlation between fragile layers and task starting layers or optimize the robustness of fragile layers based on the functional positioning of task starting layers, resulting in targeted deficiencies in robustness optimization and inability to achieve coordinated improvement of task performance and robustness \cite{IAENGhowto:liu2023robustness, IAENGhowto:hernandez2024linearity}.

\subsection{Hybrid Architecture Design}
With the diversified development of LLM architectures, hybrid combinations of new structures such as GateDeltaNet and Mamba with traditional DecoderLayers, as well as designs like MoE dynamic gating and layer routing, have become important directions to improve model efficiency and performance \cite{IAENGhowto:fedus2022, IAENGhowto:lieber2024}. These hybrid architectures balance inference efficiency and expressive ability by integrating the core advantages of different architectures, becoming the mainstream trend in current LLM architecture design, such as the hybrid design of GateDeltaNet and traditional Full Attention in Qwen3.5 and the Mamba-Transformer hybrid architecture that fuses Mamba’s fast scanning capability with Transformer’s global modeling capability \cite{IAENGhowto:qwen35github, IAENGhowto:lieber2024}. However, the current design of hybrid architectures still faces core bottlenecks: the selection of layer switching nodes and the determination of module ratios mostly rely on researchers’ empirical settings, lacking scientific theoretical basis and quantitative support \cite{IAENGhowto:zhao2023}. Existing analysis frames can only adapt to layer-wise analysis of a single architecture, unable to uniformly quantify and compare the layer-wise functions of different architectures, and existing research including non-probing methods fails to apply task starting layer positioning results to layer division and module switching of hybrid architectures, failing to establish an association system between task starting layer positioning, layer function division, and hybrid architecture design, which leads to redundancy or capability loss in hybrid architecture design and inability to give full play to the complementary advantages of various architectures \cite{IAENGhowto:lieber2024, IAENGhowto:xu2024layerwise}.

\begin{figure*}[t]
    \centering
    \includegraphics[width=\textwidth]{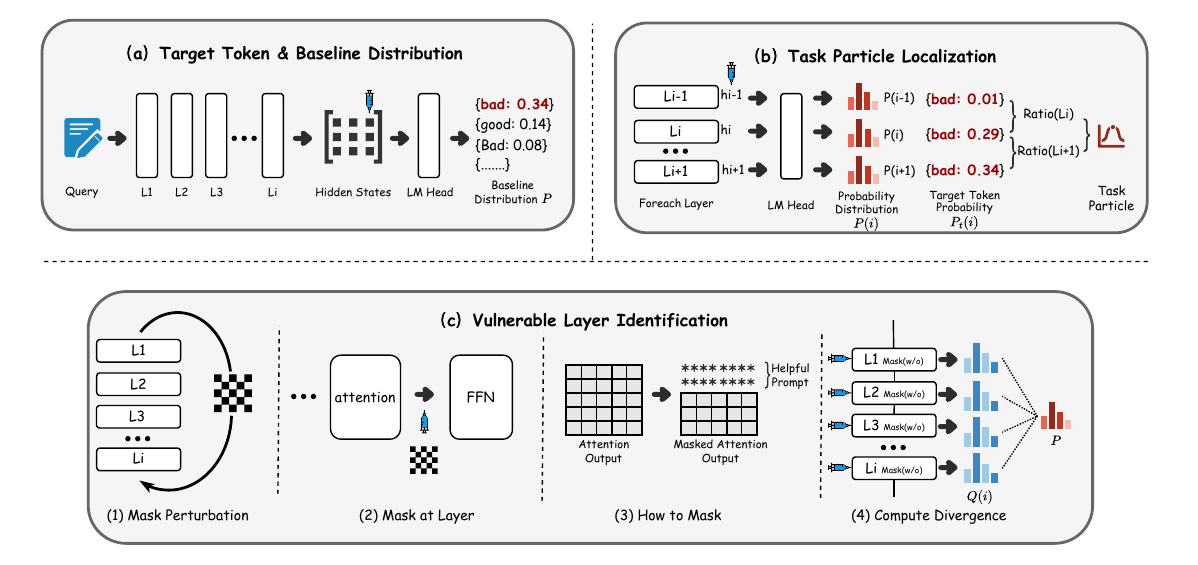}
    \caption{Overview of the LayerTracer two-phase analysis pipeline. 
    (a) A query is fed into the model to obtain the final-layer probability distribution \(P\), from which the highest-probability token is selected as the target token \(t^*\).
    (b) For each layer \(i\), we extract the hidden state \(h_i\), project it via the LM head to obtain the layer-wise probability distribution \(P(i)\), and record the target token probability \(P_t(i)\). The relative increase ratio \(\text{Ratio}(i) = (P_t(i) - P_t(i-1)) / P_t(i)\) is computed, and the layer with the maximum ratio is identified as the task particle, representing the onset of task-specific reasoning.
    (c) To simulate parameter variations during partial-layer training, we apply masking perturbations to the attention output of each layer. The divergence between the baseline distribution \(P\) and the perturbed output distribution \(Q(i)\) is measured via Jensen-Shannon divergence, and the layer with the highest divergence is marked as the vulnerable layer.
    }
    \label{fig:framework}
\end{figure*}
\section{method}
\subsection{Definitions and Quantification Formulas}
\label{sec:define}
To enable a joint quantitative analysis of the task singularity and vulnerable layers across arbitrary large model architectures, we formally define two core metrics and provide their quantification methods. These definitions serve as the theoretical foundation for the subsequent analytical pipeline.

First, this paper introduce the concept of task singularity. The task singularity is defined as follows: Let the target token be the vocabulary item with the highest probability in the model's final-layer output distribution, and let $P_t(l)$ denote its probability at layer $l$. We compute the relative probability increase ratio between consecutive layers as:
\begin{equation}
\mathrm{Ratio}(l) = \frac{P_t(l) - P_t(l-1)}{P_t(l)}
\end{equation}
where $P_t(l-1)$ represents the target token probability at the preceding layer $l-1$. The layer $l$ that maximizes $\mathrm{Ratio}(l)$ is identified as the task singularity. This definition is strictly architecture-agnostic and applicable to any hierarchical large language model. It precisely characterizes the critical onset layer where task-specific semantic encoding begins to emerge, serving as a principled boundary for module switching in hybrid architectures.

Second, this paper define the vulnerable layer. The vulnerable layer is defined as follows: Taking the final-layer output probability distribution of the unperturbed model as the reference, we apply a masking perturbation to the core processing module at layer $l$ (e.g., attention or state-space block). We then compute the Jensen-Shannon (JS) divergence between the final output distribution under perturbation and the original unperturbed distribution. The layer $l$ that yields the maximum JS divergence is designated as the vulnerable layer. This quantification relies on a general distributional discrepancy metric rather than architecture-specific components, enabling seamless adaptation to the layer-wise vulnerability analysis of arbitrary model designs. The JS divergence is formally defined as:
\begin{equation}
\mathrm{JS}(P \parallel Q) = \frac{1}{2} \mathrm{KL}(P \parallel M) + \frac{1}{2} \mathrm{KL}(Q \parallel M)
\end{equation}
where $M = \frac{1}{2}(P + Q)$ denotes the average distribution, $P$ represents the original output probability distribution, $Q$ represents the perturbed output distribution, and $\mathrm{KL}(\cdot \parallel \cdot)$ denotes the Kullback-Leibler divergence. This formulation accurately quantifies the distributional shift induced by structural perturbation. The layer exhibiting the maximal divergence is highly sensitive to interference, marking it as a critical bottleneck for model robustness optimization.

\subsection{Pipeline}
\noindent \textbf{Part I: Task Particle Localization.}
To accurately locate the task particle of LLMs with arbitrary architectures, we analyze the probability trajectory of a target token across network layers, which is identified as the token with the maximum probability in the final layer's output distribution. The entire process is built on the hierarchical hidden state extraction and probability mapping results, with Top-K fixed at 10 to ensure the consistency and pertinence of data analysis. First, we load the pre-trained model through a universal model loading interface that adapts to different architectures without modifying core code, then extract the layer-end token hidden states of each target layer, block the propagation of subsequent layers to ensure the purity of single-layer behavior analysis, and normalize the hidden states to eliminate differences caused by different architectural encoding methods. After mapping the normalized hidden states to the vocabulary probability distribution via the language model head and softmax function, we screen the Top-10 candidate words and their corresponding probabilities for each layer. Based on these probability data, we calculate the relative probability increase ratio for each layer using the quantitative formula defined in Section~\ref{sec:define}, namely the relative change in target token probability between consecutive layers. The layer with the maximum ratio is identified as the task particle, representing the critical layer where the model starts to effectively learn and execute the target task. For multi-task scenarios, we conduct aggregation analysis on the task particles of all tasks to obtain the global task particle distribution law, which provides a scientific basis for the layer division of hybrid architectures.

\noindent \textbf{Part II: Vulnerable Layer Identification.}
Vulnerable layer identification is implemented based on the hidden state data and probability mapping results obtained in the task particle localization process, taking the probability distribution of the model's last layer as the ground truth to quantify the layer's sensitivity to disturbances. The core purpose of this process is to simulate the scenario where some layers are frozen or trained during model pre-training: when layers are trained, their parameters will inevitably change, and such changes will affect the model's output stability. Therefore, we apply uniform masking interference and noise addition to the output of the attention module of each target layer to simulate the parameter changes of the trained layers, thereby accurately evaluating the sensitivity of each layer to parameter disturbances and identifying the robustness bottleneck of the model. To ensure the fairness and consistency of the interference method, the same masking intensity and noise amplitude are adopted for all layers. For each interfered layer, we calculate the JS divergence between the final output probability distribution after interference and the final output probability distribution without interference using the JS divergence formula defined in Section~\ref{sec:define}. The layer with the maximum JS divergence value is determined as the vulnerable layer, which reflects the strongest sensitivity of the layer to parameter disturbances and is the key focus of model robustness optimization.  The detailed process is detailed in Algorithm~\ref{alg:layertracer}.

\begin{algorithm}[ht]
\caption{LayerTracer Algorithm}
\label{alg:layertracer}
\begin{algorithmic}[1]
\Require Target LLM $M$, layer indices $\mathcal{L}=\{l_1,\dots,l_k\}$, prompt $p$
\Ensure Task particle $l_{\text{particle}}$, vulnerable layer $l_{\text{vuln}}$

\Statex \textbf{Phase 1: Task Particle Localization}
\State Load $M$; forward pass to obtain reference distribution $P_{\text{ref}}$ at final layer
\State Identify target token $t^* \gets \arg\max_v P_{\text{ref}}(v)$
\For{each layer $l \in \mathcal{L}$}
    \State Extract hidden state $h_l$; block propagation after $l$
    \State Compute layer distribution $P_l \gets \text{softmax}(\text{lm\_head}(h_l))$
    \State Record $p_l \gets P_l(t^*)$
\EndFor
\For{each $l \in \mathcal{L} \setminus \{l_{\min}\}$}
    \State Compute $\text{Ratio}(l) \gets \frac{p_l - p_{l-1}}{p_l}$
\EndFor
\State $l_{\text{particle}} \gets \arg\max_{l} \text{Ratio}(l)$
\Statex \textbf{Phase 2: Vulnerable Layer Identification}
\State Take $P_{\text{ref}}$ as the unperturbed baseline distribution
\For{each layer $l \in \mathcal{L}$}
    \State Apply masking perturbation to the core module output at layer $l$ \Comment{Simulate parameter update}
    \State Forward pass to obtain perturbed output distribution $Q_l$
    \State Compute divergence $D_l \gets \mathrm{JS}(P_{\text{ref}} \parallel Q_l)$
\EndFor
\State $l_{\text{vuln}} \gets \arg\max_{l} D_l$

\State \Return $l_{\text{particle}}, l_{\text{vuln}}$
\end{algorithmic}
\end{algorithm}

\section{Results and Discussion}
\label{result}
\subsection{Experimental Settings}
All experiments are conducted on a server equipped with 8 NVIDIA A800 GPUs. This study utilizes four Qwen3 variants spanning distinct parameter scales, specifically 0.6 B, 4 B, 8B, and 14B parameters, as the primary subjects for layer-wise analysis. The experiment aims to characterize the distribution trends of Task Particles and Vulnerable Layers across varying model capacities. For each layer, the hidden state is projected via the language modeling head to obtain the full vocabulary distribution, from which we retain only the top-10 candidate tokens with the highest probabilities.

\subsection{Datasets and Metrics}
The experiments utilize the AntSynNET~\cite{IAENGhowto:nguyen2017distinguishing} dataset, from which the first 500 samples are selected and sequentially partitioned into ten categories to facilitate the hierarchical feature evaluation of the Qwen3 series models. To achieve a comprehensive quantitative assessment of model layer-wise characteristics, three core metrics are designed. Specifically, the Relative Probability Increase Ratio and JS Divergence are employed to locate Task Particles and quantify single-layer vulnerability, respectively; their mathematical formulations have been detailed in Section~\ref{sec:define}. Furthermore, this study introduces the Layer-wise Relative Stability (LRS) metric to measure the uniformity of overall hierarchical robustness. The LRS is defined as:
\begin{equation}
\mathrm{LRS} = \sqrt{\frac{1}{N-1} \sum_{l=1}^{N} \left( \mathrm{JS}(l) - \overline{\mathrm{JS}} \right)^2}
\end{equation}
where $N$ denotes the total number of layers in the model, $\mathrm{JS}(l)$ represents the JS divergence value at layer $l$, and $\overline{\mathrm{JS}}$ indicates the mean JS divergence across all layers. A lower LRS value signifies smaller discrepancies in inter-layer vulnerability, implying a more stable and uniform hierarchical structure.

\begin{figure*}[t]
    \centering
    \subfloat[Qwen3-0.6B]{\label{fig:a}
        \includegraphics[width=0.32\textwidth]{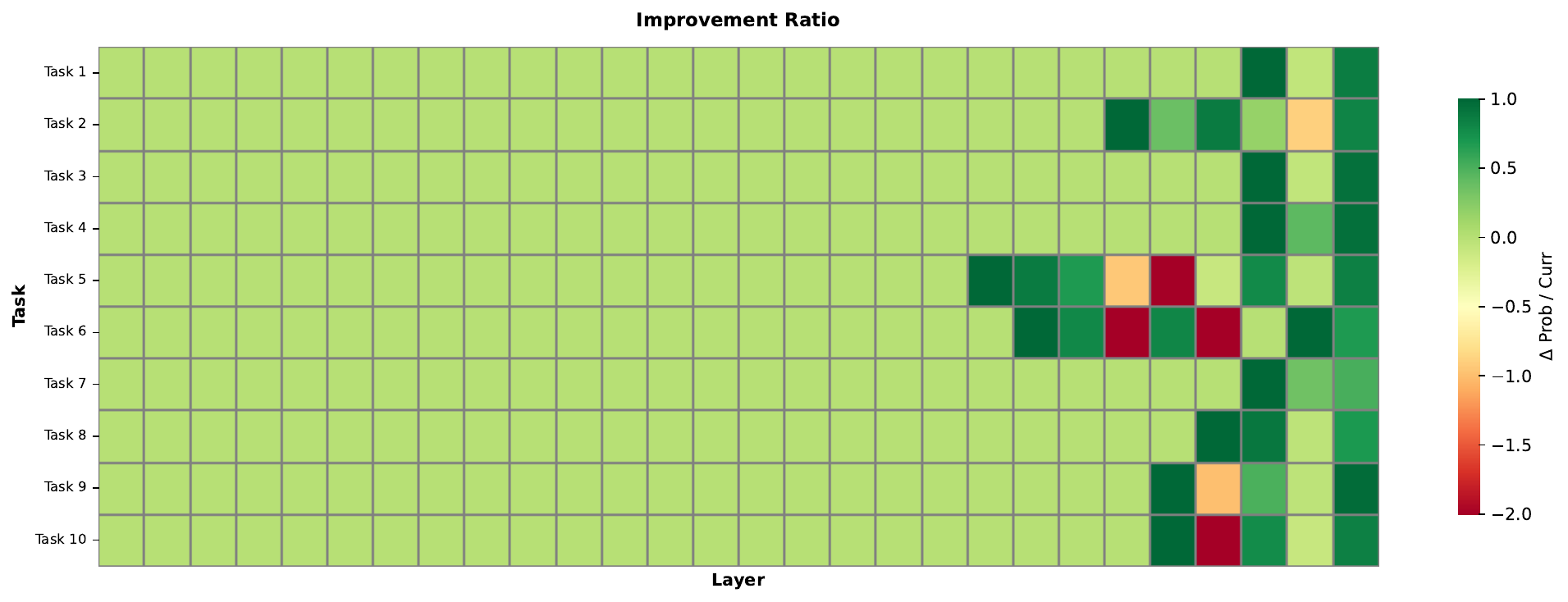}}
    \hfill
    \subfloat[Qwen3-8B]{\label{fig:b}
        \includegraphics[width=0.32\textwidth]{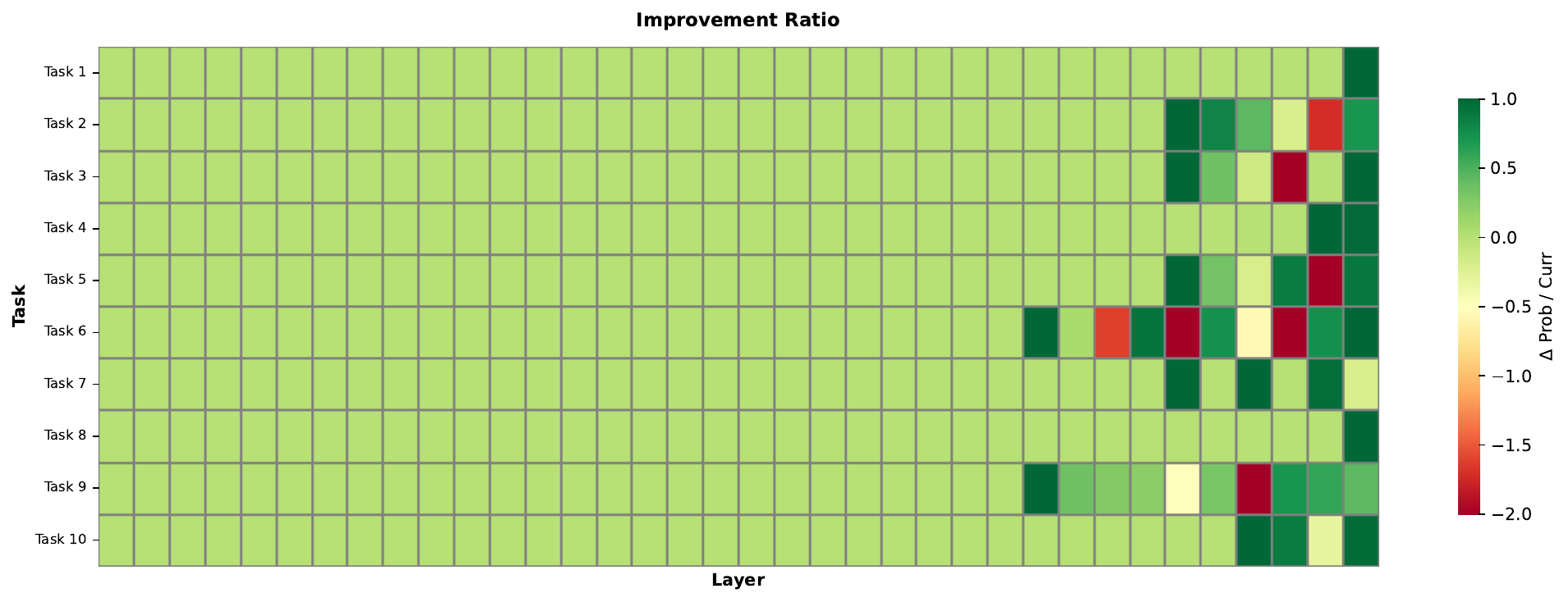}}
    \hfill
    \subfloat[Qwen3-14B]{\label{fig:c}
        \includegraphics[width=0.32\textwidth]{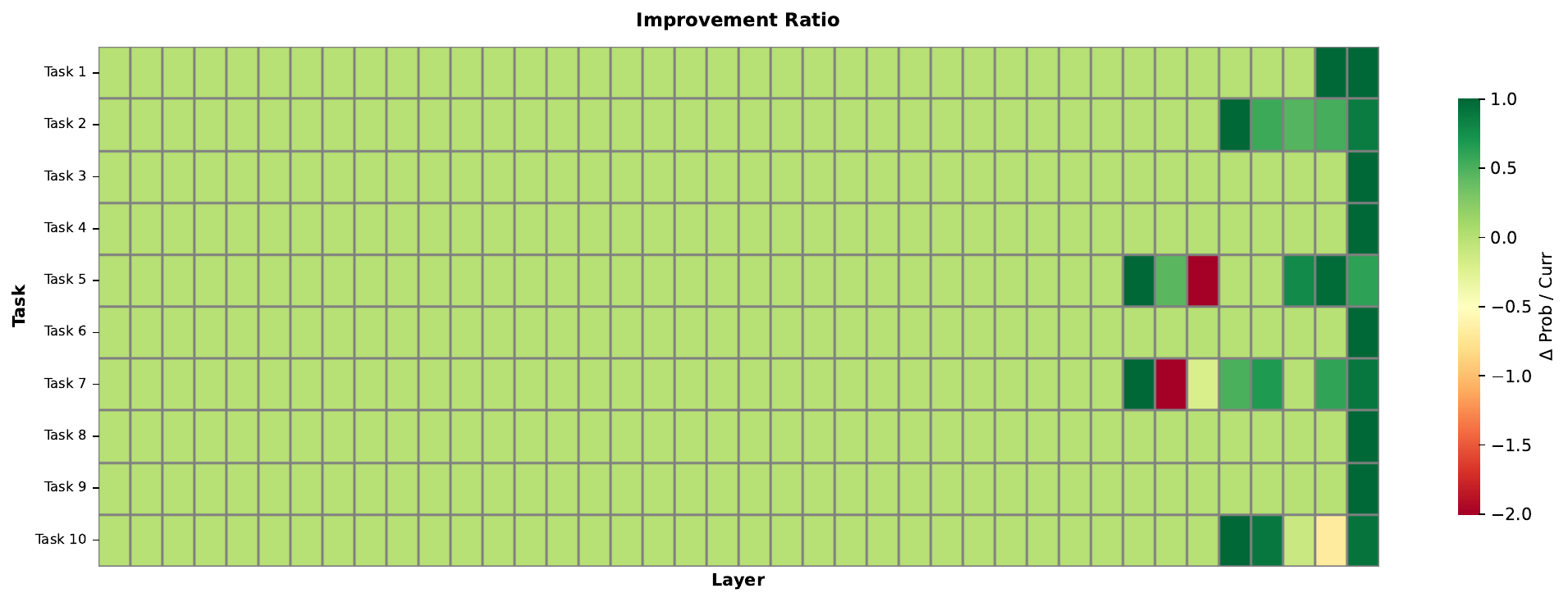}}

    \subfloat[Qwen3-0.6B]{\label{fig:d}
        \includegraphics[width=0.32\textwidth]{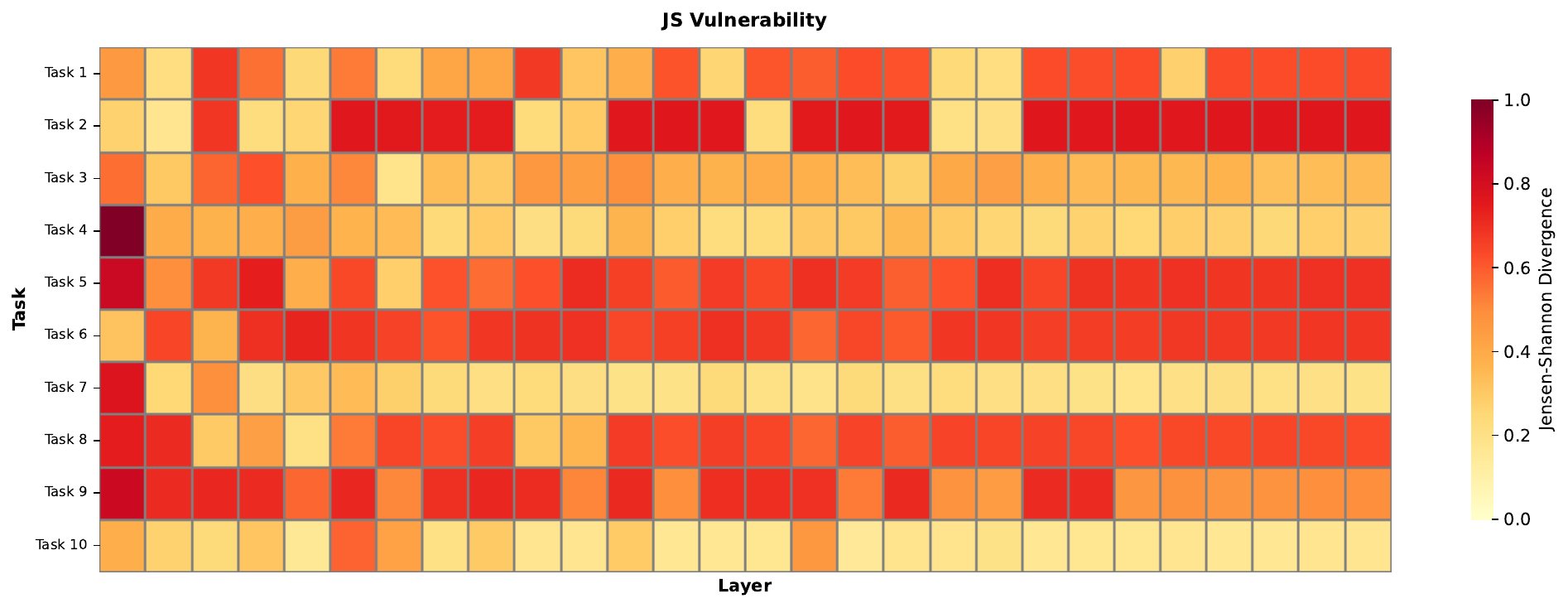}}
    \hfill
    \subfloat[Qwen3-8B]{\label{fig:e}
        \includegraphics[width=0.32\textwidth]{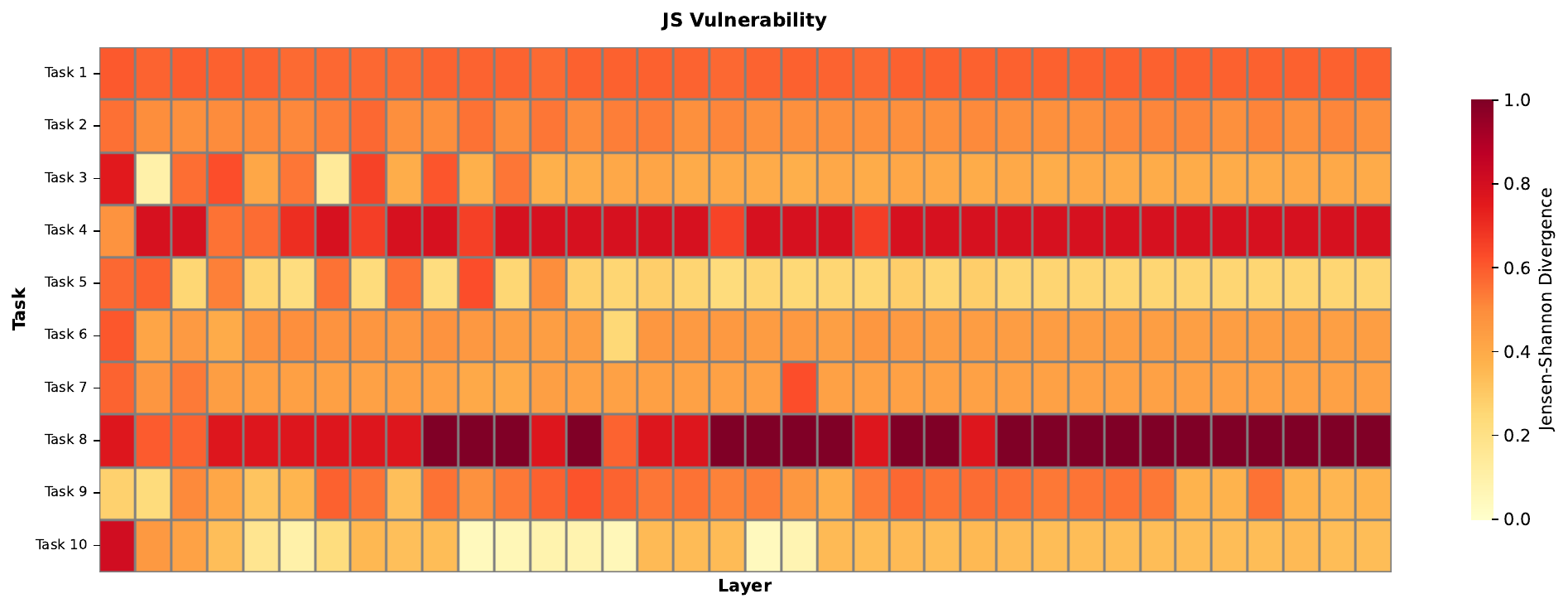}}
    \hfill
    \subfloat[Qwen3-14B]{\label{fig:f}
        \includegraphics[width=0.32\textwidth]{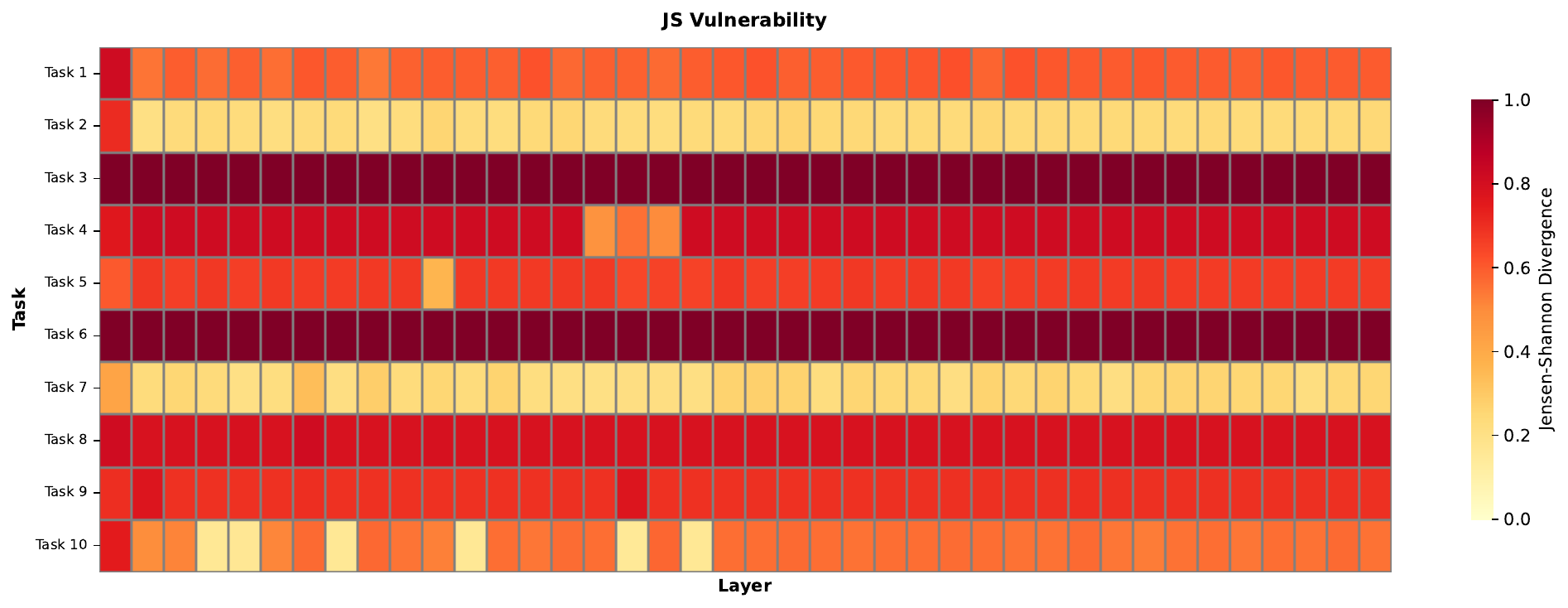}}

    \caption{Distribution details in AntSynNET dataset. (a)-(c) present the variation of Ratio across different layers for Qwen3-0.6B to Qwen3-14B. A deeper green color indicates a higher Ratio, indicating a greater relative increase in the probability of generating the Target Token at the current layer and a more significant effect of the model at this layer; a red color indicates that the current layer inhibits the generation of the Target Token. (d)-(f) present the variation of JS across different layers for Qwen3-0.6B to Qwen3-14B, representing the difference between the final output probability distribution of the model after interfering with the i-th layer and that before interference. A deeper color indicates a larger difference and higher vulnerability of the current layer of the model.
    }
    \label{fig:main_results} 
\end{figure*}
\subsection{Main Results}
\label{sec:main_results}
Figures~\ref{fig:a}--\ref{fig:c} illustrate the layer-wise evolution of the $\mathrm{Ratio}$ across the Qwen3 series (0.6B to 14B). A consistent deep-emergence phenomenon is observed across all parameter scales: the transition from negligible or negative ratios to significant positive surges consistently initiates in the latter half of the network (approximately layers $> N/2$). This sharp inflection point marks the critical depth where the model shifts from low-level feature abstraction to active task-specific semantic construction. Notably, the location of this surge exhibits a relative depth dependency; rather than being anchored to fixed absolute layer indices, the task particle consistently emerges in the relative posterior region of the architecture, suggesting that the depth required for semantic consolidation scales proportionally with the total network depth.

Figures~\ref{fig:d}--\ref{fig:f} reveal distinct patterns in hierarchical vulnerability as model capacity increases. The Qwen3-0.6B model exhibits a pronounced U-shaped vulnerability distribution, where both shallow and deep layers demonstrate markedly higher sensitivity to perturbations compared to mid-depth layers. This uneven profile corresponds to a high Layer-wise Relative Stability (LRS) value, indicating a fragile structural topology where boundary layers act as bottlenecks. Such instability implies that small-scale models are highly susceptible to semantic misalignment if hybrid architectures indiscriminately modify shallow or deep components. 

In contrast, as the parameter scale expands to 8B and 14B, the vulnerability profile becomes progressively smoother. While deep layers inherently remain sensitive due to their proximity to the output head, the extreme disparity between boundary and middle layers diminishes significantly, yielding substantially lower LRS values. This flattening trend confirms that larger models possess a more balanced robustness distribution and a more stable global topology, enabling them to effectively buffer local perturbations and maintain coherent semantic flow even under structural interference.

\section{Conclusion}

This study provides a mechanistic basis for hybrid architecture design by addressing two core questions. First, the specific hybrid ratio in models like Qwen3.5 arises because task-critical reasoning consolidates in deep layers, while boundaries remain structurally vulnerable. Anchoring high-capacity Full Attention modules in these deep regions is thus a functional necessity to ensure decision fidelity, not merely a computational trade-off. Second, we propose a depth-aware functional alignment principle for future designs: parameter-constrained models require a boundary-protection strategy, confining trainable lightweight modules to robust mid-depth zones while freezing sensitive shallow and deep layers. In contrast, larger models with flattened vulnerability profiles allow more flexible interleaving. These findings shift hybrid architecture design from heuristic trial-and-error to a quantifiable, metric-driven optimization process.


%






\ifCLASSOPTIONcaptionsoff
  \newpage
\fi


\begin{thebibliography}{1}

\bibitem{IAENGhowto:qwen2024} Qwen Team, ``Qwen2.5 technical report,'' \emph{arXiv preprint arXiv:2412.15115}, 2024.
\bibitem{IAENGhowto:qwen2tech} Qwen Team, ``Qwen2 technical report,'' \emph{arXiv preprint arXiv:2407.10671}, 2024.
\bibitem{IAENGhowto:yang2025qwen3} A. Yang, A. Li, B. Yang, et al., ``Qwen3 technical report,'' \emph{arXiv preprint arXiv:2505.09388}, 2025.
\bibitem{IAENGhowto:qwen35github} Qwen Team, ``Qwen3.5 official repository,'', 2025. 

\bibitem{IAENGhowto:openai2023} OpenAI, ``GPT-4 technical report,'' \emph{arXiv preprint arXiv:2303.08774}, 2023.
\bibitem{IAENGhowto:touvron2023} H. Touvron, T. Lavril, G. Izacard, \textit{et al.}, ``LLaMA: Open and efficient foundation language models,'' \emph{arXiv preprint arXiv:2302.13971}, 2023.
\bibitem{IAENGhowto:touvron2023llama2} H. Touvron, L. Martin, K. Stone, \textit{et al.}, ``LLaMA 2: Open foundation and fine-tuned chat models,'' \emph{arXiv preprint arXiv:2307.09288}, 2023.
\bibitem{IAENGhowto:dubey2024llama3} A. Dubey, A. Jauhri, A. Pandey, \textit{et al.}, ``The LLaMA 3 herd of models,'' \emph{arXiv preprint arXiv:2407.21783}, 2024.

\bibitem{IAENGhowto:vaswani2017} A. Vaswani, N. Shazeer, N. Parmar, \textit{et al.}, ``Attention is all you need,'' in \emph{Adv. Neural Inf. Process. Syst. (NeurIPS)}, 2017, pp. 5998--6008.
\bibitem{IAENGhowto:gu2024} A. Gu and T. Dao, ``Mamba: Linear-time sequence modeling with selective state spaces,'' in \emph{Proc. Int. Conf. Learn. Represent. (ICLR)}, 2024.
\bibitem{IAENGhowto:dao2024} T. Dao and A. Gu, ``Transformers are SSMs: Generalized models and efficient algorithms through structured state space duality,'' \emph{arXiv preprint arXiv:2405.21060}, 2024.
\bibitem{IAENGhowto:ahn2024} K. Ahn, X. Cheng, M. Song, C. Yun, A. Jadbabaie, and S. Sra, ``Linear attention is (maybe) all you need (to understand transformer optimization),'' in \emph{Proc. Int. Conf. Learn. Represent. (ICLR)}, 2024.
\bibitem{IAENGhowto:fedus2022} W. Fedus, B. Zoph, and N. Shazeer, ``Switch transformers: Scaling to trillion parameter models with simple and efficient sparsity,'' \emph{J. Mach. Learn. Res.}, vol. 23, no. 120, pp. 1--39, 2022.
\bibitem{IAENGhowto:lieber2024} A. Lieber, O. Sharir, B. Lenz, and Y. Shoham, ``Jamba: A hybrid transformer-mamba language model,'' \emph{arXiv preprint arXiv:2403.19887}, 2024.
\bibitem{IAENGhowto:gateddelta2025} S. Yang, J. Kautz, and A. Hatamizadeh, ``Gated delta networks: Improving Mamba2 with delta rule,'' in \emph{Proc. Int. Conf. Learn. Represent. (ICLR)}, 2025.

\bibitem{IAENGhowto:nguyen2017distinguishing}
K.~A. Nguyen, S. Schulte im Walde, and N.~T. Vu,
``Distinguishing antonyms and synonyms in a pattern-based neural network,''
in \emph{Proceedings of the 15th Conference of the European Chapter of the Association for Computational Linguistics: Volume 1, Long Papers},
Association for Computational Linguistics, 2017.
\bibitem{IAENGhowto:zhao2023} W. X. Zhao, K. Zhou, J. Li, \textit{et al.}, ``A survey of large language models,'' \emph{arXiv preprint arXiv:2303.18223}, 2023.
\bibitem{IAENGhowto:lin1991} J. Lin, ``Divergence measures based on the Shannon entropy,'' \emph{IEEE Trans. Inf. Theory}, vol. 37, no. 1, pp. 145--151, Jan. 1991.
\bibitem{IAENGhowto:kaplan2020scaling} J. Kaplan, S. McCandlish, T. Henighan, \textit{et al.}, ``Scaling laws for neural language models,'' \emph{arXiv preprint arXiv:2001.08361}, 2020.

\bibitem{IAENGhowto:ling2026neural}G.~Ling, Z.~Huang, Y.~Lin, J.~Li, S.~Zhong, H.~Wu, and L.~Lin,``Neural chain-of-thought search: Searching the optimal reasoning path to enhance large language models,''\emph{arXiv preprint arXiv:2601.11340}, 2026.
\bibitem{IAENGhowto:belinkov2022probing} Y. Belinkov, ``Probing classifiers: Promises, shortcomings, and advances,'' \emph{Comput. Linguist.}, vol. 48, no. 1, pp. 207--219, 2022.
\bibitem{IAENGhowto:geva2021transformer} M. Geva, R. Schuster, J. Berant, and O. Levy, ``Transformer feed-forward layers are key-value memories,'' in \emph{Proc. Conf. Empirical Methods Nat. Lang. Process. (EMNLP)}, 2021, pp. 5484--5495.
\bibitem{IAENGhowto:meng2022locating} K. Meng, D. Bau, A. Andonian, and Y. Belinkov, ``Locating and editing factual associations in GPT,'' in \emph{Adv. Neural Inf. Process. Syst. (NeurIPS)}, 2022, pp. 17\,359--17\,372.
\bibitem{IAENGhowto:gurnee2023finding} W. Gurnee, N. Nanda, M. Pauly, K. Harvey, D. Troitskii, and D. Bertsimas, ``Finding neurons in a haystack: Case studies with sparse probing,'' in \emph{Adv. Neural Inf. Process. Syst. (NeurIPS)}, 2023, pp. 56\,482--56\,501.
\bibitem{IAENGhowto:kornblith2019similarity} S. Kornblith, M. Norouzi, H. Lee, and G. Hinton, ``Similarity of neural network representations revisited,'' in \emph{Proc. Int. Conf. Mach. Learn. (ICML)}, 2019, pp. 3\,519--3\,529.
\bibitem{IAENGhowto:liu2023robustness} T. Liu, Y. Li, Q. Xie, X. Wang, and H. Li, ``A survey on the robustness of large language models,'' in \emph{Proc. Conf. Empirical Methods Nat. Lang. Process.: Findings (EMNLP Findings)}, 2023, pp. 14\,521--14\,538.
\bibitem{IAENGhowto:meng2023mass} K. Meng, D. Bau, A. Andonian, and Y. Belinkov, ``Mass editing memory in a transformer,'' in \emph{Proc. Int. Conf. Learn. Represent. (ICLR)}, 2023.
\bibitem{IAENGhowto:hernandez2024linearity} E. Hernandez, A. S. Sharma, T. Haklay, K. Meng, M. Wattenberg, J. Andreas, Y. Belinkov, and D. Bau, ``Linearity of relation decoding in transformer language models,'' in \emph{Proc. Int. Conf. Learn. Represent. (ICLR)}, 2024.
\bibitem{IAENGhowto:liu2024whatprobes} F. Liu, P. Shi, X. Liu, Y. Zhang, and G. Neubig, ``What do probes actually probe? On the role of surface statistics in linguistic probing,'' in \emph{Proc. Annu. Meeting Assoc. Comput. Linguistics (ACL)}, 2024, pp. 12\,345--12\,359.
\bibitem{IAENGhowto:chen2023layerwise} X. Chen, Y. Wang, and H. Li, ``Layer-wise probing for semantic role labeling in pre-trained language models,'' in \emph{Proc. Conf. Empirical Methods Nat. Lang. Process. (EMNLP)}, 2023, pp. 8\,901--8\,915.
\bibitem{IAENGhowto:zhao2024understanding} W. Zhao, K. Zhou, J. Li, and T. Tang, ``Understanding layer-wise representations in large language models via probing,'' in \emph{Proc. North Amer. Chapter Assoc. Comput. Linguistics (NAACL)}, 2024, pp. 2\,101--2\,115.
\bibitem{IAENGhowto:kim2023probing} S. Kim, J. Lee, and H. Park, ``Probing for factual knowledge in large language models: A layer-wise analysis,'' in \emph{Proc. Conf. Empirical Methods Nat. Lang. Process. (EMNLP)}, 2023, pp. 11\,234--11\,248.
\bibitem{IAENGhowto:xu2024layerwise} Y. Xu, Z. Li, and R. Chen, ``Layer-wise analysis of knowledge distillation in large language models,'' in \emph{Proc. Annu. Meeting Assoc. Comput. Linguistics (ACL)}, 2024, pp. 5\,678--5\,692.
\end{thebibliography}
\end{document}